\documentclass{article}

\usepackage{microtype}
\usepackage{graphicx}
\usepackage{subfigure}
\usepackage{lipsum}
\usepackage{amsmath}
\usepackage{amsfonts}
\usepackage{booktabs} 
\usepackage{mathtools}

\usepackage{hyperref}

\usepackage[accepted]{icml2019}

\icmltitlerunning{Sparse Sinkhorn Attention}

\begin{document}

\twocolumn[
\icmltitle{Sparse Sinkhorn Attention}




\begin{icmlauthorlist}
\icmlauthor{Yi Tay}{google}
\icmlauthor{Dara Bahri}{google}
\icmlauthor{Liu Yang}{google}
\icmlauthor{Donald Metzler}{google}
\icmlauthor{Da-Cheng Juan}{google}
\end{icmlauthorlist}

\icmlaffiliation{google}{Google AI}

\icmlcorrespondingauthor{Yi Tay}{yitay@google.com}

\icmlkeywords{Machine Learning, ICML}

\vskip 0.3in
]



\printAffiliationsAndNotice{} 

\begin{abstract}
We propose Sparse Sinkhorn Attention, a new efficient and sparse method for learning to attend. Our method is based on differentiable sorting of internal representations. Concretely, we introduce a meta sorting network that learns to generate latent permutations over sequences. Given sorted sequences, we are then able to compute quasi-global attention with only local windows, improving the memory efficiency of the attention module. To this end, we propose new algorithmic innovations such as Causal Sinkhorn Balancing and SortCut, a dynamic sequence truncation method for tailoring Sinkhorn Attention for encoding and/or decoding purposes. Via extensive experiments on algorithmic seq2seq sorting, language modeling, pixel-wise image generation, document classification and natural language inference, we demonstrate that our memory efficient Sinkhorn Attention method is competitive with vanilla attention and consistently outperforms recently proposed efficient Transformer models such as Sparse Transformers.
\end{abstract}

\section{Introduction}
 Learning sparse and efficient attention mechanisms has recently garnered considerable interest \cite{child2019generating,kitaev2020reformer}. While existing state-of-the-art attention models have typically relied on dense, fully-connected attention graphs \cite{vaswani2017attention}, these methods are often sub-optimal for two key reasons. First, large memory costs are incurred due to the quadratic complexity at the attention layer. Second, soft dense attention may suffer when $\ell$, the sequence length, is large and noisy. Hence, at times, sparse attentive outputs that are reminiscent of hard attention methods, may serve as a desirable inductive bias \cite{xu2015show}.

This paper proposes a new method for (1) reducing the memory complexity of the dot-product attention mechanism and (2) learning sparse attention outputs. Our method is based on a novel idea of differentiable sorting of internal representations \textit{within} the self-attention module. Our method, which we call Sparse Sinkhorn Attention, incorporates a meta sorting network that learns to re    arrange and sort input sequences. With this new sorted sequence, attention computation is reduced substantially even when considering computation only within the local neighborhood, emulating a global effect even with solely local computation of context windows.

Our method is comprised of (1) a parameterized meta sorting network $S$ for dynamically generating block-wise permutation matrices and (2) a standard local attention module that receives block-wise permuted input sequences for attention computation. Concretely, at the heart of our sorting network lives a differentiable Sinkhorn balancing mechanism \cite{adams2011ranking,mena2018learning}, which normalizes permutation matrices to belong to the Birkhoff polytope, the set of doubly stochastic matrices \cite{sinkhorn1964relationship}. 

As such, given the block-sorted input sequences, the local attention module is able to compute attention weights beyond the default local neighborhood without incurring additional computation costs. Extensive experimental results across a potpourri of language, vision and arithmetic tasks demonstrate that Sparse Sinkhorn Attention outperforms strong baselines such as standard local attention and sparse attention Transformers \cite{child2019generating}.

Notably, our proposed method is general purpose in nature and is applicable to sequence encoding, sequence decoding or seq2seq tasks \cite{sutskever2014sequence}. In order to adapt Sinkhorn balancing to decoding tasks, we propose a causal variant, i.e., Causal Sinkhorn Balancing. Moreover, for further improvement to encoding efficiency, we propose an additional \textsc{SortCut} variant of our proposed method, which dynamically truncates sequences in a data-driven manner based on a user-defined budget hyperparameter. Finally, we propose a Mixture model between the Sparse Sinkhorn Attention and standard vanilla attention, leading to further performance improvements.


Our method reduces the memory complexity from $O(\ell^2)$ to $O(B^2 + N_B^2)$ where $B=\frac{\ell}{N_{B}}$. When $\ell$ is large, this factorization of sequence length brings about substantial savings in terms of memory complexity\footnote{As an illustration, when $\ell=1024$ and $N_B=64$, this results in a memory saving factor of 240 times.}. Our \textsc{SortCut} variant further reduces complexity to linear-time, i.e.,  $O(\ell N_k)$ where $N_k$ is a user defined budget hyperparameter and $N_k <<< \ell$.

We also equip state-of-the-art Transformer models with our proposed Sparse Sinkhorn Attention, evaluating Sinkhorn Transformers on several large-scale sequence modeling tasks including language modeling on the One Billion Word Corpus \cite{chelba2013one}, pixel-wise image generation and document classification. Our proposed Sinkhorn attention remains competitive to the dense fully-connected attention while outperforming local attention and Sparse Transformers. While differentiable neural-based sorting has demonstrated some proof-of-concept promise \cite{mena2018learning}, this work demonstrates the first successful application in real large-scale problems.

To summarize, the contributions of this paper are as follows:
\begin{itemize}
    \item We propose Sparse Sinkhorn Attention, a new attention method based on dynamic, learnable sorting of internal representations. Our method is based on differentiable Sinkhorn balancing and is the first successful application of differentiable sorting on large-scale tasks.
    \item We also propose (1) Causal Sinkhorn balancing for autoregressive sequence decoding and (2) a new \textsc{SortCut} encoding scheme that further improves encoding efficiency by dynamically truncating sequences during attention computation.
    \item Our proposed methods reduce the memory complexity of dot-product attention while remaining competitive with or outperforming dense vanilla attention.
    \item We conduct extensive experiments on large-scale generative modeling tasks. On all tasks, Sinkhorn Transformers match and/or outperform vanilla Transformers while consistently outperforming Sparse Transformers \cite{child2019generating} and Local Attention Transformers. \end{itemize}

\section{Related Work}
A natural and intuitive yet naive method typically employed for efficiently learning attention involves using a fixed window size. This method, usually referred to as local attention \cite{luong2015effective}, has served as a simple and quick fix to run attention models on long sequences. An obvious weakness is that tokens in a window do not have access to context outside the window, restricting the expressiveness and its capability to model long-term dependencies. The study of window (or block-based) local attention has also been an emerging field of research \cite{shen2018bi,tay2019simple,qiu2020blockwise,child2019generating,parmar2018image}.

Building upon the notion of local windows, Sparse Transformer \cite{child2019generating} proposed factorizing the attention computation into local and strided operations, delegating different heads to focus on different sparse patterns. They demonstrate promising results, establishing Sparse Transformer as one of the canonical methods\footnote{That said, Sparse Attention requires highly specialized GPU kernels for efficient computation. This generally makes the approach less appealing, e.g., for portability purposes such as running on TPU pods.} for efficient attention computation. 

While our method also relies on sequence partitioning, we note that there have been several orthogonal but related efforts. Reformer \cite{kitaev2020reformer}, proposes locality sensitive hashing as a means to reduce the memory complexity of self-attention. Transformer-XL \cite{dai2019transformer} adopts recurrence to cache hidden states across long sequences, which spurred further interest in modeling and compression of long term dependencies \cite{rae2020compressive}. Star Transformer \cite{guo2019star} performs attention sparsification by converting the dense graph into a star\-shaped topology using a shared relay node. However, while this method enables linear-time complexity, its setup makes it difficult for causal masking, making the Star Transformer useful only for encoding.

Learning sparse outputs in attention models has also garnered reasonable interest. The key idea behind sparse weights (i.e., hard attention) is that they enable the model to only focus on a limited number of items at a time \cite{xu2015show, shen2018reinforced}. This can be a useful inductive bias when the input sequence is long and/or noisy, serving as a denoising filter. Moreover, hard attention can also improve inference speeds, as demonstrated by methods such as Sparsemax \cite{martins2016softmax}. Along a similar vein, this is also reminiscent of Sparse Mixture of Experts \cite{shazeer2017outrageously}, which performs a sparse selection of outputs (experts) for prediction tasks.

Our proposed method is not only a new way of learning efficient attention but also a new way of sparsification. At the core of our approach lies a Sinkhorn ranking operation \cite{adams2011ranking} that is used for learning differentiable rankings over internal representations. Leveraging the Gumbel reparameterization trick \cite{jang2016categorical}, Gumbel Sinkhorn Networks \cite{mena2018learning} proposed stochastic maximization over the set of possible latent permutations. The core novelty of our work lies in the introduction of neural sorting as a means to sparsify and improve the efficiency of well-established attention networks.

\section{Sparse Sinkhorn Attention}

In this section, we introduce our proposed Sparse Sinkhorn Attention and provide a high-level overview. In our method, the input sequence $X$ of length $\ell$ is partitioned into $N_b$ blocks in which each block has a length of $b$ tokens. Notably, the original idea of block-based local attention is to allow tokens to only attend to tokens within the same block. However, this restricts the global receptive field and limits the ability for local attention models to model long term dependencies. 

\begin{figure}[H]
    \centering
    \includegraphics[width=1.0\linewidth]{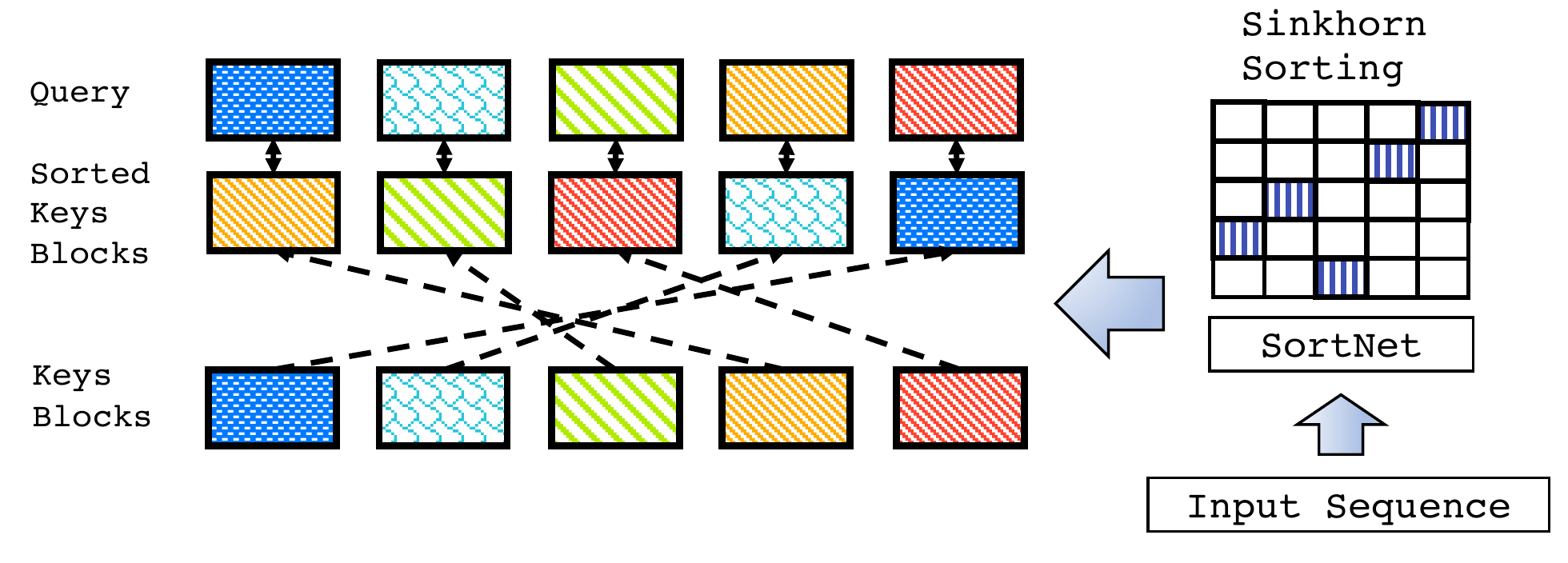}
    \vspace{-2em}
    \caption{Overview of Sparse Sinkhorn Attention. A Meta Sorting Network learns to sort sequences to enable efficient quasi-global local attention.}
    \label{fig:my_label}
\end{figure}

Our proposed method mitigates this problem by neural sorting of blocks and receptive fields (neighborhoods). More concretely, instead of attending to tokens in the same block, each token attends to tokens in the newly \textit{sorted} block, which may actually be far apart in the original unsorted sequence. Sorting blocks instead of individual tokens is also more intuitive, since we do not wish to break connections between nearby tokens, i.e., it would be reasonable for each token to still maintain an approximate neighborhood.

\subsection{Learning to Sort} In order to learn to sort, we introduce a Sorting Network (SortNet) for learning relaxed permutation matrices. Since our sorting function is differentiable, the parameters of the SortNet are also trained together in an end-to-end fashion. The SortNet accepts an input sequence of $\ell$ vectors of $d$ dimensions and partitions them into blocks.
\begin{equation}
X' = \psi_P(X)    
\end{equation}
The function $\psi_P(.)$ is a blockwise pooling operation that maps $\mathbb{R}^{\ell \times d} \rightarrow \mathbb{R}^{N_B \times d}$ and $X' \in \mathbb{R}^{N_B \times d}$. In SortNet, we adopt:
\begin{equation}
\psi_P(X)_{i}=\sum^{(i+1) * \ell_B}_{j=i * \ell_B}(X_j)
\end{equation}
which is equivalent to taking the sum of embeddings of all tokens belonging to the local window.
Our trainable SortNet is defined as follows:
\begin{equation}
R_{i} = P(X'_i)
\end{equation}
  where $i$ refers to the block index.  $P(.)$ is an arbitrary parameterized function which accepts an input vector of $d$ dimensions and returns a vector of $N_B$ dimensions. For example, we may parameterize $P(X)$ using a two layered feed-forward network with ReLU activations.
\begin{equation}
P(X) =  \sigma(W_{B}\sigma(W_{P}(X) + b_{P}) + b_{B}
\end{equation} 
where $W_{P} \in \mathbb{R}^{d \times d}$ and $W_{B} \in \mathbb{R}^{d \times \ell_{B}}$. Essentially, the key idea is that each block learns a projection to $N_B$ other blocks, effectively learning the position that it is supposed to be shifted (or permuted) to. 

\subsubsection{Sinkhorn Normalization}
The matrix $R$ becomes a sorting matrix (or permutation matrix) if it is doubly stochastic (matrix is nonnegative and both rows and columns all sum to 1). More specifically, a permutation matrix is special case of a doubly stochastic matrix (where rows and columns sum to 1 and all entries are either $0$ or $1$). Since every permutation matrix is a convex combination of doubly stochastic matrices, we consider learning doubly stochastic matrices as a a form of relaxed permutation matrix.

We consecutively normalize the rows and columns of the sorting matrix $R$, i.e., a process of Sinkhorn normalization \cite{adams2011ranking}. Here, the number of iterations $N_k$ is a user defined hyperparameter. This procedure is described as follows:
\begin{align*}
S^{0}(R) &= \text{exp}(R)  \\ 
S^{k}(R) &= F_{c}(F_{r}(S^{k-1}(R))) \\
S(R) &= \lim_{k \rightarrow \infty} S^K(R)
\end{align*}
where $F_{r}, F_{c}$ are the row and column wise normalization function defined as follows:
\begin{align*}
F^k_{c}(X) &= F^{k-1}_{c}(X) \oslash (X1_{\ell}1^{\top}_{N}) \\
F^k_{r}(X) &= F^{k-1}_{r}(X) \oslash (1_{\ell}1^{\top}_{N}X) \\
\end{align*}
where $\oslash$ is the element-wise division operator, $N$ is the length of the input matrix and $\textbf{1}$ is a vector of ones. In practice, we perform calculations in log domain for improved stability. 
\begin{align*}
F^k_{c}(X) &= F^{k-1}_{c}(X) - \log (\exp (X1_{\ell})1^{\top}_{N}) \\
F^k_{r}(X) &= F^{k-1}_{r}(X)  - \log (1_{\ell}1^{\top}_{N}\exp (X)) \\
\end{align*} 
To this end, \cite{sinkhorn1964relationship} shows that iterative normalization of $R$ converges to the doubly stochastic limit if $R$ has support, i.e., a nonnegative matrix with a positive diagonal. Note that since $R$ is nonnegative by design due to the usage of ReLU in $P(X)$. Gradients of the iterative Sinkhorn normalization can be computed, enabling end-to-end training.
\subsubsection{Neural Sorting of Sequences}
The generated permutation matrix is then used to sort the input sequence. This is described by a simple matrix multiplication of $R$ against the blocked input sequence $X'$:
\begin{align*}
X_S = U({R}B(X))  
\end{align*}
where $B(.)$ converts input sequence into block-wise representations, i.e., $X' \in \mathbb{R}^{N_B \times (B \times d)}$ and $U(.)$ converts the block-wise sequences back into token-wise sequences. $U(.)$ and $B(.)$ can be interpreted as block-wise reshaping operators. Since $R$ is doubly stochastic, multiplying a partitioned sequence by $R$ is equivalent to sorting it. 

\subsection{Sparse Sinkhorn Attention}
The key idea of the Sparse Sinkhorn Attention is to operate on block sorted sequences. Hence, the revised computation for the attention mechanism can now be written as:
\begin{align*}
    A_{ij} = 
    \begin{cases}
    (Q_{i}\psi_S(K)_{j}^\top) + Q_{i}(K)_{j}^\top),& \text{if } \lfloor{{j}/{\ell}}\rfloor = \lfloor{i/{\ell}}\rfloor\\
    0              & \text{otherwise}
\end{cases}
\end{align*}
$\psi(.)$ is the neural sorting function. Intuitively, this is identical to only enabling attention without a certain local neighborhood, albeit with key values sorted in a block-wise fashion. Subsequently, to compute and soft-select from the value matrix, we compute:
\begin{align*}
Y = \text{Softmax}(A)\psi_S(V)
\end{align*}
Here, the value matrix is also sorted accordingly. In practice, we share the sorting operator between the key and values. The secondary term $Q_{i}(K)_{j}^\top)$ is the standard local attention which is added to the mixture. In practice, attention weights are only computed when $\lfloor{{j}/{\ell}}\rfloor = \lfloor{i/{\ell}}\rfloor$.

\subsubsection{Gumbel Noise}
For $S(X)$ to approximate the doubly-stochastic permutation matrix, we leverage the Gumbel categorical reparameterization trick \cite{jang2016categorical}. Concretely, we inject Gumbel noise into our sorting operator:
\begin{align*}
S(X) = S(\frac{(X + \epsilon)}{\tau})    
\end{align*}
where $\epsilon$ is the injected standard i.i.d Gumbel noise and $\tau$ is the temperature hyperparameter. Intuitively, lowering the temperature brings $S(X)$ to be closer to a permutation matrix with discrete $1$s and $0$s.

\subsubsection{Multihead Sparse Sinkhorn Attention} 
We have previously described the computation of a single Sinkhorn attention head. Similar to dot product attention, utilizing the multi-headed variation is straightforward. 
\begin{align*}
Y_{G} = F_{H}([Y_{1} \cdots Y_{N_H}])    
\end{align*}
where $Y_{i}$ is the output of the $i$-th attention head. $F_{H}$ is a linear transform layer with kernels $W \in \mathbb{R}^{(N_{H} \times d) \times d}$. Notably, our implementation learns a sorting network on a per head basis, i.e., we do not share the same permutation matrix $R$ across all heads.

\subsubsection{Mixture Model}
Finally, we also consider a variant where the Sinkhorn Attention is used to model an alternate view of the input sequence. Concretely, we leverage the combination of the Sinkhorn attention by mixing it with the vanilla standard dot product attention.
\begin{align*}
Y = \text{Softmax}(A)\psi_S(V) + \text{Softmax}(QK^\top)V
\end{align*}
Notably, the mixture mode regresses to the same quadratic complexity of vanilla self-attention. However, we hypothesize that the side network may provide an alternative and diverse view, ultimately improving performance.

\subsection{Causal Sparse Sinkhorn Attention}
Our Sinkhorn Attention not only involves sorting sequences but also learning the sorting order in a content-based fashion. To this end, pertaining to learning causal attention (i.e., no information from the future should leak to the present) there are two cases that we have to be careful about. The first is that current time steps should never have access to future time steps. Hence, if block $i$ is sorted into a new position $p<i$, then it is being masked out. This produces an inductive bias that favors sorting orders that produce sorting between nearby blocks.
\subsubsection{Causal Sorting Networks}
The second case is the content-based and dynamic learning of sorting networks. To maintain the causal property, it would not be plausible to generate permutation matrices based on global information such as the sum of tokens in a sequence. Hence, the matrix $R$ is generated using the cumulative sum of embeddings instead. This is described as follows:
\begin{equation}
\psi_P(X)_{i}=\sum^{(i* \ell_B+1)}_{j=0}(X_j) \:\:\:\text{and}\:\:\:
R= P(\psi_P(X))
\end{equation}
Since our attention operates based on the idea of blocks, we use the first token in the block as its representative embedding. The cumulative sum operator allows the model to learn a permutation matrix conditioned on all previous context information leading up to the current block. 

\subsubsection{Causal Sinkhorn Balancing}
We note that the original Sinkhorn balancing requires knowledge of the future tokens for normalization. For causal self-attention, this is undesirable and non-permissible. Hence, we develop a causal variation of the typical Sinkhorn Balancing method which performs masking of the future while performing iterative normalization. 
\begin{align*}
F^{k}_{c}(X) &= F^{k-1}_{c}(X) - \log (\exp (M(X)1_{\ell})1^{\top}_{N}) \\
F^{k}_{r}(X) &= F^{k-1}_{r}(X) - \log (1_{\ell} 1^{\top}_{N} M(\exp (X))) \\
\end{align*}
where $M(.)$ is a masking function. 
\begin{equation}
 M(x)= 
\begin{cases}
    1,& \text{if } j\geq i\\
    0,              & \text{otherwise}
\end{cases}
\end{equation}

\subsubsection{Connections to Learnable Sparsity}
Due to the computation of causal Sinkhorn, it is expected that some blocks may be masked out, i.e., a block is masked out if it is sorted into an earlier position (i.e, $i'<i$). Essentially, the Sorting Network is also learning which tokens to mask by determining the sorting sequence.


\subsection{\textsc{SortCut} Sinkhorn Attention}
We propose an additional variant of our Sparse Sinkhorn Attention which we call \textsc{SortCut}. In this method, we propose a post-sorting truncation of the input sequence, essentially performing a hard top-k operation on the input sequence blocks within the computational graph. While most attention models mainly re-weight or assign near-zero weights during training, our method enables us to explicitly and dynamically truncate the input sequence. Specifically,
\begin{align*}
Y = Softmax(Q\psi_S(K)_{[:n]}^\top)\psi_S(V)_{[:n]}
\end{align*}
where $n$ is the \textsc{SortCut} budget hyperparameter. A caveat is that this mode may only be performed on the the Transformer encoder unless self-attention is explicitly computed again for every time-step in autoregressive decoding. 
\begin{figure}[H]
    \centering
    \includegraphics[width=1.0\linewidth]{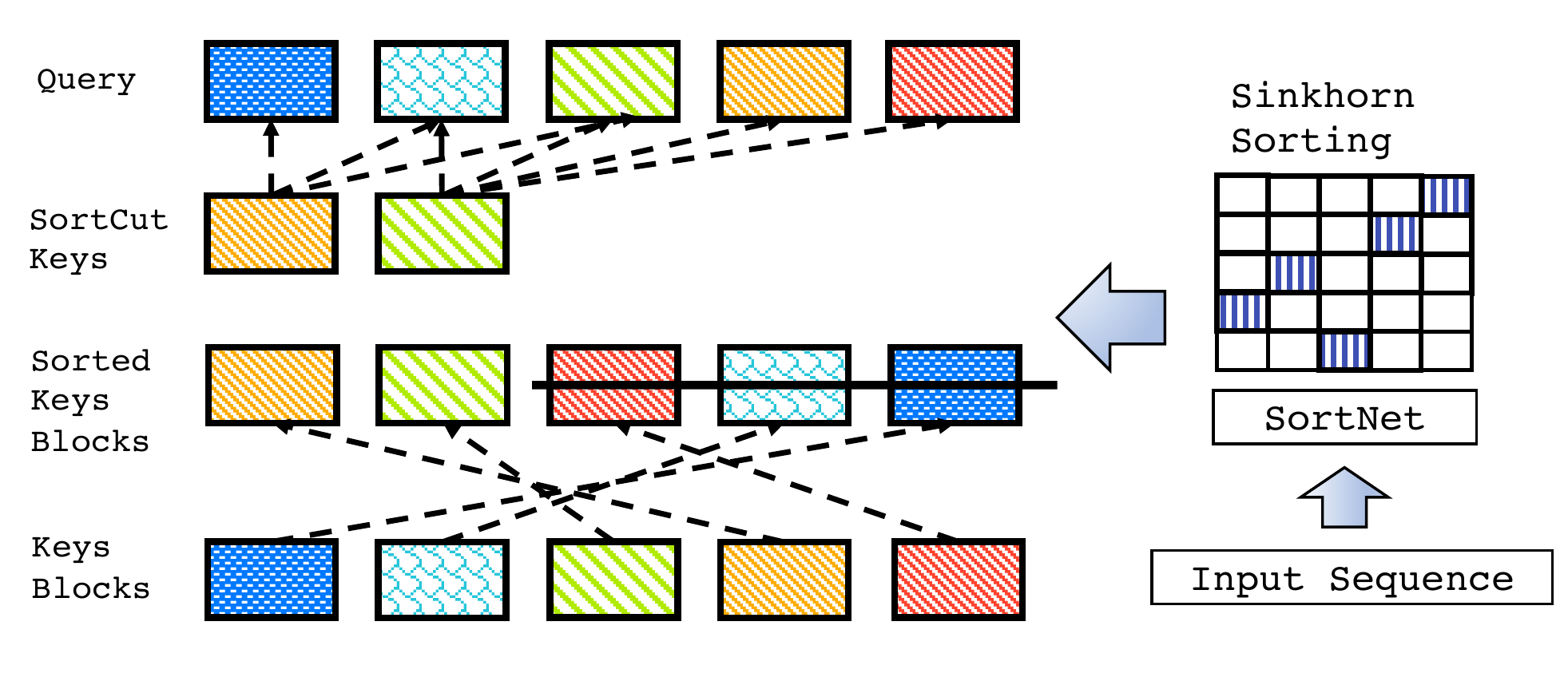}
    \vspace{-2em}
    \caption{Overview of the proposed SortCut Encoding Scheme.}
    \label{fig:my_label2}
\end{figure}
\section{Complexity Analysis}
The vanilla Transformer has a self-attention memory complexity of $O(\ell^2)$ where $\ell$ is the input sequence length. Our proposed Sinkhorn model reduces this to $O(B^2 + (\frac{\ell}{N_{B}})^2)$ where $B=\frac{\ell}{N_{B}}$. Essentially, this is equivalent to the memory complexity of local attention models. The \textsc{Sortcut} Sinkhorn encoder has a memory complexity of $O(\ell N_k + ({N_{B}})^2)$ where $N_k$ is the budget hyperparameter. Since $\frac{\ell}{\ell_B}<<\ell$, the complexity of the \textsc{SortCut} encoder can be reduced to $O(\ell)$.

\section{Experiments}
We evaluate the effectiveness of our proposed method for five tasks, including algorithmic sorting, language modeling, pixel-wise image generation, document classification and natural language inference. All our experiments are run on the open source Tensor2Tensor framework \cite{tensor2tensor}. If not stated otherwise, our Sinkhorn Transformers adopt the following global hyperparameters - temperature $\tau$ tuned among $\{0.25,0.50,0.75,1.0\}$, number of sort iterations tuned among $\{2,5,10,20\}$. Block size is largely dependent on the maximum length of the problem domain. Our tasks are designed to capture a wide and diverse range of scenarios such as medium to long range (e.g., 256 to 2048) and also covering a range of encoding focused and/or decoding focused tasks. 

\subsection{Algorithmic Tasks}
We first test our model on a toy algorithmic sorting task. The task is cast as a sequence transduction problem (seq2seq) where the model is tasked to output a sorted sequence of an integer sequence.
\paragraph{Experimental Setup} 
We use the \texttt{algorithmic\_sort\_problem} task in Tensor2Tensor. In this task, we train our models to sort sequences of $\ell=256$ and evaluate on sequences of length $2\ell$ (i.e., 512) to probe for generalization ability and ensure the models are not just simply memorizing. We evaluate based on exact match (EM) and edit distance (the lower the better). The exact match metric is defined by the number of test sequences that the model gets entirely correct. The dataset consists of $100K$ train examples and $1000$ test examples. For Sinkhorn Transformers, we utilize Sparse Sinkhorn Attention for both encoding and decoding. We train all models for $200$k steps using the default Transformer base hyperparameter. We compare against vanilla Transformers, local attention Transformers and Sparse Transformers.

\begin{table}[]
    \centering
    \begin{tabular}{ccc}
    \hline
        	Model & Edit Dist. &	EM \\
        	\hline
   
Transformer &	0.4252 &	45.69 \\
Local Attention ($32$) & 0.4340	& 21.12 \\
Sparse Transformer ($32$) & 0.4176 & 46.88 \\
\hline
Sinkhorn Transformer ($8$) & 0.4156	& 43.65 \\ 
Sinkhorn Transformer ($16$) & 0.4071 &	48.23 \\
Sinkhorn Transformer ($32$) &	\textbf{0.4054} &	\textbf{49.24} \\
\hline
    \end{tabular}
    \caption{Evaluation on Algorithmic Sequences of length $256$.}
    \label{tab:sorting}
\end{table}

\paragraph{Results on Sorting Task} Table \ref{tab:sorting} reports our results on the algorithmic sorting task. Sinkhorn Transformers outperform all other Transformer variants, including the vanilla attention model. Sparse Transformer outperforms dense attention, which demonstrates the usefulness of sparse inductive biases. The local attention performs the worst, which demonstrates that some extent of global knowledge is required to solve this task. 

\subsection{Language Modeling}
We evaluate on the LM1B (Language Modeling One Billion) dataset \cite{chelba2013one}, a large-scale language modeling benchmark. We evaluate on subword-level and character level language modeling on this task. 
\paragraph{Experimental Setup}
We implement our model in the Tensor2Tensor framework, using the packed TPU setting. Tokens are split into $32$k word pieces and sentences are shuffled. For word level language modeling, we use the default Tensor2Tensor hyperparameters\footnote{\url{tensor2tensor/models/research/lm_experiments.py}}. Concretely, we evaluate two model sizes, \textsc{base} and \textsc{big} corresponding to \texttt{lmx\_base} and \texttt{lmx\_h2k\_f8k} respectively. All models are trained for $300K$ steps on 16 TPU V2 Chips. For Sinkhorn Transformers and the corresponding Local Attention baseline, we tune the block size $B \in \{16,32,64\}$. We compare against the vanilla Transformer baseline, Local Attention Transformers, and Sparse Transformers \cite{child2019generating}. We implement Sparse Transformers in Tensor2Tensor by referencing the original open source code using the fixed attention scheme \cite{child2019generating}. However, instead of integrating the specialized cuda kernels, we manually simulated masking to achieve an equivalent implementation. We use a block length of $N_B=64$ and fixed stride of $c=8$. For character-level language modeling, owing to the overall larger sequence length, we use a maximum sequence length of $1024$ and use a fixed block length of $128$.

\paragraph{Results on Subword level Language Modeling} 
Table \ref{tab:word_lm} reports results on subword level language modeling. On both parameter settings, Sinkhorn Transformers outperform all local attention models and Sparse Transformer. Pertaining to relative performance between Sinkhorn and local attention, the performance gain at each $B$ setting ranges from $2-3$ perplexity points. Notably, on the base setting, Sinkhorn Transformer outperforms the vanilla Transformer at $B=32$ and $B=64$. At $B=16$, Sinkhorn Transformers remain competitive to base Transformers. On the big setting, Sinkhorn Transformers fail to outperform vanilla Transformers, but still perform reasonably well despite being more memory efficient. Finally, the Sinkhorn Mixture model outperforms all models.  
\begin{table}[]
    \centering
    \begin{tabular}{ccc}
    \hline
    & \multicolumn{2}{c}{Perplexity} \\
        Model & Base & Big \\
        \hline
        Transformer \cite{vaswani2017attention} &	41.57 & 27.59\\
        \hline
Local Attention ($16$) & 	44.62  & 30.14\\
Local Attention ($32$) & 	44.23  & 29.32\\
Local Attention ($64$) &	44.23 & 28.97\\
Sparse Transformer ($64$) & 41.89 & 28.77  \\ 
\hline
Sinkhorn Transformer ($16$) & 	42.64 & 29.42\\
Sinkhorn Transformer ($32$) & 	41.29  & 28.48\\
Sinkhorn Transformer ($64$) &	40.79 &
28.39 \\ 
Sinkhorn Mixture & \textbf{40.11} & \textbf{27.34} \\
\hline
    \end{tabular}
    \caption{Experimental results on Language Model One Billion (LM1B) benchmark using the Base (50M parameter) and Big (430M) setting.}
    \label{tab:word_lm}
\end{table}
\paragraph{Results on Character-level Language Modeling} Table \ref{tab:char_lm} reports our experimental results (bytes per char) on character level language modeling. On both settings (base/big), our proposed Sinkhorn Transformer outperforms both local attention and Sparse Transformer, which affirms its effectiveness as an efficient attention method. On the contrary, local attention performs substantially worse compared to its counterparts, likely due to not having much global context. From this set of experiments, the vanilla full attention Transformer outperforms all efficient attention methods. However, our Sinkhorn Mixture model outperforms the Transformer baseline, achieving the best performance for both parameterizations.

\paragraph{Comparison with the State-of-the-art} Table \ref{tab:sota} reports our best scores relative to the state-of-the-art\footnote{To the best of our knowledge, \cite{shazeer2018mesh} is the best performing model on per-word perplexity. \cite{baevski2018adaptive} and \cite{dai2019transformer} report per-token perplexity}. Notably, our best performing Sinkhorn Transformer remains competitive with the High Budget MoE \cite{shazeer2017outrageously} and Evolved Transformer \cite{so2019evolved} models. This demonstrates the overall competitiveness of Sinkhorn Transformers. Unfortunately, we were unable to outperform Mesh Tensorflow \cite{shazeer2018mesh} on our setup, which consists of $5$ billion parameters. Nevertheless, we consider our results to be reasonable given the improved memory complexity.

\begin{table}[]
    \centering
    \begin{tabular}{ccc}
    \hline
    Model & \# Params & Perplexity \\
    \hline
        Low Budget MoE  & 5.0B &	34.10 \\
Transformer (Big) & 141M &	30.44 \\
Evolved Transformer (Big) & 151M  & 	28.60 \\
High Budget MoE	& 5.0B & 28.00 \\
Mesh Tensorflow & 4.9B & \textbf{24.00} \\ 
\hline
Sinkhorn Transformer & 450M & 28.39 \\
Sinkhorn Transformer & 1.9B & 27.34 \\
\hline
\end{tabular}
    \caption{Comparison with other published works on LM1B that uses per-word Perplexity. Sinkhorn Transformer remains competitive to other Transformer models and High Budget MoE models.}
    \label{tab:sota}
\end{table}

\begin{table}[]
    \centering
    \begin{tabular}{ccc}
    \hline
         & \multicolumn{2}{c}{Bytes per char (Bpc)} \\
        Model & Base & Big \\
        \hline
Local Attention	& 2.559 & 1.825 \\ 
Transformer & 	1.283 & 1.121 \\
Sparse Transformer	& 1.300 & 1.134 \\
\hline
Sinkhorn Transformer &	1.295 & 1.132 \\
Sinkhorn Mixture & \textbf{1.270}& \textbf{1.119} \\
\hline
    \end{tabular}
    \caption{Experimental results on character level language modeling on LM1B with sequence lengths of $1024$ characters.}
    \label{tab:char_lm}
\end{table}

\subsection{Pixel-wise Image Generation} 
This section introduces and reports results on pixel-wise image generation task. This task models unconditional generative modeling of images by modeling images as flat sequences of pixels.
\paragraph{Experimental Setup} We evaluate our model on pixel-by-pixel image generation using the Tensor2Tensor framework. We use the CIFAR-10 dataset. Similar to language modeling, we evaluate using bytes per dimension (Bpd), a common metric for evaluating generative modeling of images. In this task, images are flatted to sequences of $3076$ bits which probes for long-term sequence modeling capabilities. We train all models using the base parameterization for $500K$ steps with a batch size of $1$.
\begin{table}[H]
    \centering
    \begin{tabular}{cc}
    \hline
        Model  & Bpd \\
        \hline
Local Attention	& 4.200 \\ 
Transformer \cite{vaswani2017attention} & 	3.198   \\
Sparse Transformer ($256$)	& 3.227   \\
\hline
Sinkhorn Transformer ($256$) &	\textbf{3.197}  \\
Sinkhorn Mixture & 3.199 \\
\hline
    \end{tabular}
    \caption{Experimental results on pixel-wise image generation (CIFAR-10)}
    \label{tab:imagegen}
\end{table}

\paragraph{Results on Image Generation}
Table \ref{tab:imagegen} reports our results on the pixel-wise image generation task. Our proposed Sinkhorn Transformer outperforms all baselines. The local attention model performs the worst, which can be intuitively attributed to lack of global knowledge. While keeping the local window identical, our model also outperforms Sparse Transformer which demonstrates its utility as an efficient attention method. Finally, the Sinkhorn Mixture performs worse than the ordinary Sinkhorn Transformer, suggesting that a restricted (and learned) global view may serve as a useful inductive bias. 

\subsection{Text Classification}
We evaluate several text classification benchmarks from the Tensor2Tensor framework. These tasks are mainly encoding only tasks, which allows us to benchmark the \textsc{Sortcut} encoding scheme. 

\paragraph{Experimental Setup} We experiment on both sentiment analysis and natural language inference. For the former, we use the standard open source IMDb sentiment \cite{maas2011learning} and Sentiment Treebank (SST) dataset \cite{socher2013recursive}. For the latter, we use two natural language inference (NLI) datasets, i.e., Stanford NLI \cite{bowman2015large} and MultiNLI \cite{williams2017broad}. 

For sentiment analysis, we evaluate on both character and word level. We set the maximum length of tokens to be 512/2048 for word/character level tasks respectively. We implement our models using Tensor2Tensor using the \textsc{tiny} Transformer setting (2 layers). Hyperparameters between our Sinkhorn Transformer and the vanilla Transformer remains identical. Token embeddings are initialized randomly. We train our models for $15000$ steps for IMDb and SST and $500000$ steps for NLI tasks. For all experiments, we use a batch size of $4096$ tokens per batch. Models are trained with a single V100 GPU.

For natural language inference, experimental setup follows the Tensor2Tensor setup where premise and hypothesis are concatenated into one long input sequence. Word embeddings are also randomly initialized. We use the Transformer tiny hyperparameter for this task. We would like to emphasize that our experimental setup for these tasks differs from the standard usage of these datasets.
\begin{table}[]
    \centering
    \begin{tabular}{ccccc}
    \hline
   &  \multicolumn{2}{c}{IMDb} &  \multicolumn{2}{c}{SST} \\
        Model  &	Word &	Char & Word & Char\\
         \hline
\cite{vaswani2017attention} &	\textbf{85.12} &	62.77 & 76.83 & 57.45
\\
\hline
Sinkhorn ($8$) &	82.51	 &   63.78 & 74.08& \textbf{62.27} \\ 
Sinkhorn  ($16$)	& 82.00	& 62.05 & 76.15 & 56.08  \\
Sinkhorn ($32$)&	83.54	& 62.87 & \textbf{77.52}& 58.14 \\
\hline
\textsc{Sortcut} ($2$x$8$) &84.32
& 64.53 & 73.85 & 56.65 \\
\textsc{Sortcut} ($2$x$16$) & 80.12 & \textbf{64.87}& 74.31& 58.14  \\
\textsc{Sortcut} ($2$x$32$) &84.43  &62.80 & 75.81 & 56.42 \\
\hline
    \end{tabular}
    \caption{Experimental results on word and character level document classification on IMDb dataset and SST datasets.}
    \label{tab:sentiment}
\end{table}

\begin{table}[]
    \centering
    \begin{tabular}{ccc}
    \hline
         Model & SNLI & MNLI   \\
         \hline
        Transformers \cite{vaswani2017attention} & 78.87 & 53.69\\ 
        \hline
        Sinkhorn ($8$) & 68.34 & 52.15 \\
        Sinkhorn ($16$) & 77.77  & 52.09\\
         Sinkhorn ($32$) & 78.62  & 54.25\\
        \hline
         Sortcut Sinkhorn ($2$x$8$) & 75.84 &  48.88 \\
          Sortcut Sinkhorn ($2$x$16$) & \textbf{80.30} & 49.78  \\
           Sortcut Sinkhorn ($2$x$32$) &  79.39 & \textbf{55.80} \\
           \hline
    \end{tabular}
    \caption{Experimental results on natural language inference.}
    \label{tab:nli}
\end{table}
\paragraph{Results on Sentiment Analysis} Table \ref{tab:sentiment} reports results on sentiment analysis. Sinkhorn Transformers demonstrate promising results on sentiment analysis datasets on both word and character level. Even with significant memory savings, Sinkhorn Transformers are able to outperform or remain competitive with the baseline Transformer model. We also take this chance to benchmark the \textsc{Sortcut} encoder, which further reduces memory complexity. On all settings, we find that the \textsc{Sortcut} variation can achieve similar performance to not only the standard Sinkhorn Transformer but also the vanilla Transformer.

\paragraph{Results on Natural Language Inference}
Table \ref{tab:nli} reports results on SNLI and MNLI tasks. We find that both Sinkhorn and Sortcut Sinkhorn are able to outperform the vanilla Transformer. This task demonstrates the effectiveness of the SortCut variant despite the improvement of memory complexity over the standard Sinkhorn Transformer.

\section{Analysis}
In this section, we study the effect of certain modeling choices. 
\begin{table}[H]
    \centering
    \begin{tabular}{lc}
    \hline
    Modeling Choice & Perplexity \\ 
    \hline
   (1) $P(X)=\sigma(F_2(\sigma(F_1(X))))$ & 41.70 \\
       (2) $P(X)=F_2(\sigma(F_1(X)))$  &  41.38  \\
      (3) $P(X)=\sigma(F_1(X))$   & 41.34  \\
       (4) $P(X)=F_1(X)$ & 41.29 \\
         (5) $K=V$ & 42.26 \\
         (6) $N_k$=0 (no sinkhorn) & 52.40  \\
         \hline
    \end{tabular}
    \caption{Effect of different Sorting Network variants on $b=32$ on LM1B (lower is better). $F(.)$ refers to linear transformation layers.}
    \label{tab:model_choice}
\end{table}

\subsection{Effect of Modeling Choices} We are mainly interested in the effects of varying the Sorting Network model. Table \ref{tab:model_choice} reports ablation studies on various model configurations. In (1) to (4), we vary the sorting network model. In (5), we experiment with a scheme to tie the weights of $K$ and $V$ (this is because they share the same permutation matrix). From Table \ref{tab:model_choice}, the best model for learning the sorting matrix is a linear layer, which signifies that the sorting network can be a simple model. We also observed that, more often than not, sharing the key-values seem to hurt performance. Finally in (6), we set $N_k=0$ which is equivalent to not performing Sinkhorn normalization on $R$. We observe that performance degrades substantially and performs the worse of all ablative variations. 

\subsection{Hard or Soft Sorting?}
Figure \ref{fig:temperature} reports the effect of varying Sinkhorn balancing temperatures. Keeping all other variables constant, we varied the temperature of the Gumbel Sinkhorn balancing mechanism. Overall, we find that maintaining a high temperature (inclined towards soft sorting) works better than a more discrete (hard) form of sorting. On this task, the optimal temperature is at $\tau=0.75$.


\subsection{Effect of Sorting Iterations}
\begin{figure}[H]
\begin{minipage}{0.48\linewidth}
  \centering
   \centering
    \includegraphics[width=1.0\linewidth]{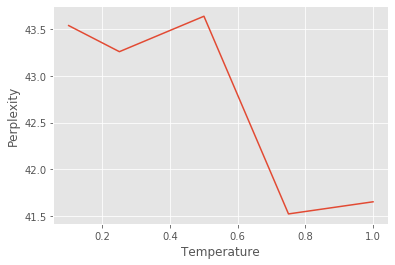}
    \caption{Effect of $\tau$ on perplexity scores (LM1B).}
    \label{fig:temperature}
\end{minipage}\hfill
\begin{minipage}{0.48\linewidth}
  \centering
  \vspace{1em}
     \includegraphics[width=1.0\linewidth]{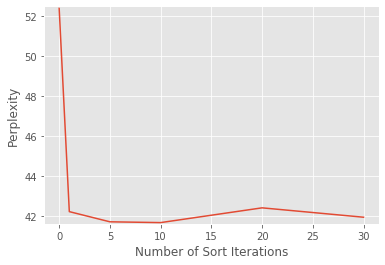}
    \caption{Effect of sorting iterations $k$ on perplexity scores (LM1B).}
    \label{fig:sortiter}
\end{minipage}
\label{fig:analysis}
\end{figure}

Figure \ref{fig:sortiter} reports the performance trend with respect to $N_k$, the number of sorting iterations. Overall, a small number of sorting iterations is sufficient for good performance. No sorting at all performs extremely bad while  the optimal number of sorting iterations seems to be $5-10$. Conversely, increasing the number of sorting iterations (beyond $20$) seem to hurt perplexity scores. 

\section{Conclusion}
We proposed Sparse Sinkhorn Attention, a new efficient and sparse method for attention computation. Our work demonstrates the utility of neural sorting of internal representations within the attention module on a multitude of large-scale generative modeling and classification tasks. On these benchamrks, our proposed Sinkhorn Transformer outperforms or remains competitive to vanilla Transformer and sparse Transformer models on a multitude of applications while being memory efficient.
\bibliography{icml2019}
\bibliographystyle{icml2019}





\end{document}